\documentclass[lettersize,journal]{IEEEtran}
\usepackage{amsmath,amsfonts}
\usepackage{algorithmic}
\usepackage{algorithm}
\usepackage{array}
\usepackage[caption=false,font=normalsize,textfont=sf]{subfig}
\usepackage{textcomp}
\usepackage{stfloats}
\usepackage{url}
\usepackage{verbatim}
\usepackage{graphicx}
\usepackage{cite}
\usepackage{enumitem}
\begin{document}

\title{\huge TIMRL: A Novel Meta-Reinforcement Learning Framework for Non-Stationary and Multi-Task Environments}

\author{Chenyang Qi, Huiping Li,~\IEEEmembership{Senior Member,~IEEE}, and Panfeng Huang,~\IEEEmembership{Senior Member,~IEEE}
        \thanks{Chenyang Qi, Huiping Li are with the School of Marine
        Science and Technology, Northwestern Polytechnical University, Xi'an
        710072, China (e-mail: cy\_q@mail.nwpu.edu.cn; lihuiping@nwpu.edu.cn).}
        \thanks{Panfeng Huang is with the School of Astronautics, Northwestern Polytechnical University, Xi'an
        710072, China (e-mail: pfhuang@nwpu.edu.cn).}
        \thanks{This work has been submitted to the IEEE for possible publication. 
        Copyright may be transferred without notice, after which this version may no longer be accessible.}
        }

\maketitle

\begin{abstract}
In recent years, meta-reinforcement learning (meta-RL) algorithm has been proposed to improve sample efficiency in the field of decision-making and control, 
enabling agents to learn new knowledge from a small number of samples.
However, most research uses the Gaussian distribution to extract task representation, which is poorly adapted to tasks that change in non-stationary environment.
To address this problem, we propose a novel meta-reinforcement learning method by leveraging Gaussian mixture model 
and the transformer network to construct task inference model.
The Gaussian mixture model is utilized to extend the task representation and conduct explicit encoding of tasks.
Specifically, the classification of tasks is encoded through transformer network to determine the Gaussian component corresponding to the task.
By leveraging task labels, the transformer network is trained using supervised learning.
We validate our method on MuJoCo benchmarks with non-stationary and multi-task environments.
Experimental results demonstrate that the proposed method dramatically improves sample efficiency and accurately recognizes the classification of the tasks, 
while performing excellently in the environment.

\end{abstract}


\section{Introduction}
\IEEEPARstart{R}{einforcement} learning (RL) has achieved remarkable achievements in solving complex and challenging tasks\cite{ref31}, \cite{ref32}.
However, training an excellent agent using RL algorithms typically requires a significant amount of interaction with the environment.
Agents, unlike humans, cannot learn from previous tasks and acquire experience that allows them to quickly master new tasks with minimal attempts.
The topic of how to enable agents to leverage prior task experience to rapidly adapt to new tasks is both valuable and urgent. 

Meta-reinforcement learning (meta-RL) aims to improve the efficiency and generalization of RL in the face of new tasks through a meta-learning perspective. 
The existing research on meta-RL can generally be summarized into three categories.
The first one is a gradient-based method that generates policies for new tasks by optimizing the meta-policy 
and adjusts the meta-parameters online during the meta-learning\cite{ref1},\cite{ref2}. 
The second one is recurrence-based method, 
aiming to leverage the advantage of recurrent neural networks (RNNs) in handling sequential data to assist in constructing the meta-RL framework\cite{ref3},\cite{ref4}.
All the above methods use the on-policy method for updating during meta-learning, which greatly reduces the training efficiency.
As a representative of the third method, 
the context-based algorithm PEARL\cite{ref5} is proposed to improve the performance of the algorithm by using the off-policy approach.
PEARL combines context-based task inference with the off-policy RL algorithm SAC\cite{ref6} and decouples task inference and global policy learning.

In stationary and single-task environment, traditional meta-RL algorithms have demonstrated excellent performance in adapting to new tasks.
For example, the mobile robot is able to adapt to the specified changing running speed after training.
However, if there are obstacles on the road, the robot need learn to jump or turn at the same time.
When dealing with this scenario, traditional algorithms may perform poorly, 
and they struggle to effectively handle the requirements of adapting to rapidly changing speeds and avoiding obstacles.
The traditional context-based meta-RL framework learns the task representation from historical experience, 
using Gaussian distribution as a representation of the latent space, which results in limitations in representing non-stationary and multi-task environments.
Therefore, our emphasis is on the task inference process, which is capable of utilizing a small amount of task experience to accurately identify the classification of tasks 
and map them to their corresponding latent spaces.

For the above purpose, we introduce task inference with GMM and transformer for efficient Meta-RL (TIMRL), an context-based algorithm that can efficiently adapt to non-stationary and multi-task environments.
We adopt a GMM-based framework in our task inference model for meta-RL, replacing the single Gaussian distribution, 
and design the loss function for agent during meta-training to adapt the task environment more efficiently and accomplish faster exploration. 
At the same time, we use independent module to recognize the classification of tasks and design the recognition network by transformer network, which can achieve efficient processing of task data sequences.
Before the recognition network, pre-processing is introduced to normalize the state-action dimensional spaces in a multi-task environment.
To ensure the accuracy of recognition, supervised learning is employed in the model for training.
This approach allows us to model each classification of task as a separate Gaussian component, 
thus enhancing the model's ability to handle complex tasks, especially in multi-task environments.
In order to improve the efficiency of training and accurate of recognition, we decouple the GMM framework and recognition network during training. 

To evaluate our algorithm, we extend the MuJoCo benchmarks to the non-stationary environment based on \cite{ref7} 
and design the multi-task environment to validate the adaptability of our algorithm. 
Experimental results show that our algorithm exhibits excellent performance in terms of training efficiency and asymptotic performance, especially in multi-task environments.
The main contributions can be summarized as follows.

\begin{itemize}[leftmargin=2.5em]
\item[1)] 
A novel and efficient Meta-RL algorithm (TIMRL) that significantly impoves the adaptability in non-stationary and multi-task environments is developed. 
We utilize the framework of GMM to construct task inference model and design the recognition network based on transformer according to task distribution.
Following the above approach, we formulate our context-based meta-RL framework where the respective optimization objectives in each phases are proposed.
\item[2)]
We decouple the GMM framework and recognition network in train steps to improve the efficiency of training and the accuracy of classification for the recognition,
while the supervised learning method is introduced to train the recognition network.
\end{itemize}

\section{related work}

\subsection{Meta Reinforcement Learning}
The research on meta-RL can be mainly divided into three categories: the gradient-based method, the recurrence-based method and the context-based method.

The representative achievement of gradient-based method is the Model-Agnostic Meta-Learning (MAML)\cite{ref2}, 
which learns to acquire initial “highly adaptive” parameters
and employs the policy gradient to quickly adjust initial parameters in order to adapt to new tasks. 
The advantage of the MAML algorithm lies in its model-agnostic, which means it can be applied to any model that can be trained using gradient descent methods.
Based on MAML, subsequent studies had conducted further explorations in terms of model-agnostic and rapid adaptability\cite{ref1},\cite{ref8},\cite{ref10}.
In addition, Al-Shedivat \textit{et al.}\cite{ref11} proposed a gradient-based meta-RL algorithm to improve the efficiency of adaptation 
in non-stationary and competitive environments.

The main feature of recurrence-based method is that the policy network of the algorithm consists of the recurrent networks\cite{ref28}.
The recurrence-based method uses prior experience to train recurrent networks to equip the network with task representations, 
thereby enabling agent to quickly adapt to new tasks\cite{ref12},\cite{ref29}.
RL$^2$\cite{ref3} utilized RNNs to receive and process time-series data and encoded information using the network weights to represent task-related information.
Mishra \textit{et al.}\cite{ref4} proposed a lightweight meta-learner model, simple neural attentIve learner (SNAIL), 
that utilized temporal convolutions to process information from prior experiences.

The purpose of the context-based method is to train a superior task-conditioned policy.
This method uses context to derive task representation, which is provided as auxiliary input to the policy network.
PEARL\cite{ref5}, as an influential context-based algorithm based on SAC\cite{ref6}, 
adopted off-policy method that decoupled task inference from policy learning, greatly enhancing sample efficiency.
Wu \textit{et al.}\cite{ref13} decomposed the task representation and encoded different aspects of the task. 
Xu \textit{et al.}\cite{ref14}proposed Posterior Sampling Bayesian Lifelong In-Context Reinforcement Learning (PSBL),
which performs well when there are significant differences between the distributions of training and testing tasks.

\subsection{Task Inference and Task Embedding for Meta-RL}
The task inference model utilizes context to infer task features or attributes as task embeddings, 
which serves to guide the learning process for new tasks, playing a significant role in meta-RL\cite{ref30}.
We focus on those methods that based on task inference and task embedding.

Bing \textit{et al.} proposed Continuous Environment Meta-Reinforcement Learning (CEMRL) in\cite{ref7} for non-stationary environments, 
where the training strategy and inference model can achieve excellent representation and task embedding.
Jiang \textit{et al.}\cite{ref15} designed two independent exploration goals in the action and task embedding spaces, 
reconstructing optimization objectives for the inference and policy networks to enhance exploration efficiency.
A conditional variant autoencoder is designed by Chien \textit{et al.}\cite{ref16} to learn task embeddings, which enhances the generalization of the strategy.
Ada \textit{et al.}\cite{ref17} proposed Unsupervised Meta-Testing with Conditional Neural Processes (UMCNP), 
which combines parameterized policy gradient and task inference-based method.
To solve the problem of sparse rewards, Jiang \textit{et al.}\cite{ref18} proposed a Doubly Robust augmented Transfer (DRaT) method. 
Lee \textit{et al.}\cite{ref19} proposed Subtask Decomposition and Virtual Training (SDVT) to improve the performance of the algorithm in non-parametric tasks.
Lan \textit{et al.}\cite{ref20} optimized the task encoder to generate specific task embeddings, 
while learning a shared policy conditioned on the task embeddings.
These works\cite{ref21},\cite{ref22},\cite{ref23} also make an excellent contribution to improving the sample efficiency and adaptability of the algorithm.

\section{Preliminaries}

\subsection{Context-based Meta-RL}
In the framework of context-based meta-RL algorithms, agent obtains task-relevant information from context
and uses the inference model to derive an embedding code for task representation, ensuring that the agent completes the task quickly.
Specifically, agent integrates context into the latent variable space by task inference model, enabling the conditioned policy to quickly adapt to new tasks. 

As a classical context-based meta-RL algorithm, PEARL\cite{ref5} has performed excellently in various benchmarks.
It introduces the meta-learning approach based on the SAC algorithm, in which variational auto-encoder (VAE) is employed to train task inference method\cite{ref25}.
The encoder in the VAE encodes the transition $(s, a, r, s')$ to obtain independent Gaussian factors $\Psi_\phi(\mathbf{z}|\mathbf{c}_n)$. 
The product of these Gaussian factors is used to represent the prediction of the posterior $q_\phi(z|c_{1:N})$,
which is the permutation-invariant function
\begin{equation}
  \label{deqn_ex1a}
  q_\phi(\mathbf{z}|\mathbf{c}_{1:N})
  \propto\Pi_{n=1}^N\Psi_\phi(\mathbf{z}|\mathbf{c}_n)
\end{equation}
The objective of the optimization is:
\begin{equation}
  \label{deqn_ex1a}
  \mathbb{E}_{\mathcal{T}}\left[\mathbb{E}_{\boldsymbol{z}\sim q_{\boldsymbol{\phi}}(\boldsymbol{z}|\boldsymbol{c}^{\mathcal{T}})}[R(\mathcal{T},z)]+\beta
  \mathbb{K}\mathbb{L}\left(q_{\boldsymbol{\phi}}(\boldsymbol{z}|\boldsymbol{c}^{\mathcal{T}})||p(\boldsymbol{z})\right)\right]
\end{equation}where $R(\mathcal{T},z)$ represents the objective functions, $p(z)$ is a unit Gaussian prior over tasks.

\subsection{Gaussian Mixture Model}
The standard GMM is modeled based on multiple Gaussian distributions. 
In the initial stage of the task, the parameters of each Gaussian distribution (i.e. mean and variance) are unknown. 
Furthermore, the proportion of each Gaussian distribution within the overall GMM model is randomly assigned.
\begin{equation}
  \label{deqn_ex1a}
  \begin{aligned}M_{GMM}&=\lambda_1N(\mu_1,\sigma_1)+\lambda_2N(\mu_2,\sigma_2)+\ldots\\&=\Sigma\lambda_kN(\mu_k,\sigma_k)\end{aligned}
\end{equation}where $N(\mu_k,\sigma_k)$ is the Gaussian distribution, which has the parameters of mean $\mu_k$ and variance $\sigma_k^2$, $\lambda_k$ is the proportion of each Gaussian distribution.

Given N samples and K Gaussian components, we aim to determine Gaussian component each sample belongs to and estimate the parameters of each component.
The Expectation-Maximization algorithm is used to calculate the parameters of the GMM.
First, let $\gamma(ik)$ denote the probability that the $i$-th sample belongs to the $k$-th Gaussian distribution:
\begin{equation}
  \label{deqn_ex2a}
\gamma(ik)=\frac{\lambda_kN(x_i|\mu_k,\sigma_k)}{\sum_{j=1}^K\lambda_jN(x_i|\mu_j,\sigma_j)}
\end{equation}where $x_i$ is the $i$-th sample.
At the beginning each parameter in $(\ref{deqn_ex2a})$ is obtained by initializing and $\gamma(ik)$ is calculated based on the parameters.
This means that the current estimates of the model parameters are used to calculate the probabilities.
Then, the probabilities calculated above are used to update the estimates of the model parameters.
Specifically, the parameters are updated in each step:
\begin{equation}
  \label{deqn_ex1a}
    \begin{split}
      N_k=\sum_{i=1}^N&\gamma(ik), \lambda_k=\frac{N_k}N, \mu_k=\frac{1}{N_k}\sum_i^N\gamma(ik)x_i\\
      &\sigma_k^2=\frac{1}{N_k}\sum_i^N(\gamma(ik)x_i-\mu_k)^2
    \end{split}
\end{equation}where $N_k$ is the number of samples in $k$-th component.

\subsection{Transformer Network}
Transformer network was first proposed by Vaswani \textit{et al.}\cite{ref26} as an architecture to efficiently model sequences, 
which is constructed based on the self-attention mechanism.
In RL, each transition is treated as a sequence unit and the context is recognized using the efficient sequence processing capability of transformer\cite{ref27}.
The encoder in the transformer is mainly responsible for transforming the sequence of input into a fixed-length vector of representation, 
while the decoder decodes this vector into the sequence of output.
The encoder block is a stack of multiple encoder layers, and each encoder layer contains the Multi-Head attention and the Feed-Forward network.
The Multi-Head attention is one of core mechanisms in transformer, enhancing the focus on different features.
It represents each position in the input sequence as query ($Q$), key ($K$), and value ($V$), 
and computes the attention distribution between each position and other positions to obtain a weighted sum, which captures the relevant information of each position.
The equation of attention as in $(\ref{deqn_exa})$:
\begin{equation}
  \label{deqn_exa}
  \text{Attention}(Q,K,V)=\text{softmax}(\frac{QK^T}{\sqrt{d_k}})V
\end{equation}where $d_k$ is the dimension of $K$.
The equation of Multi-Head attention as in $(\ref{deqn_ex7a})$:
\begin{equation}
  \label{deqn_ex7a}
  \begin{split}
  \mathrm{MultiHead}(Q,K,V)&=\mathrm{Concat}(\mathrm{h_1},\mathrm{h_2},\dots,\mathrm{h}_i)W^O\\
  \mathrm{h}_i=\mathrm{Attention}&(QW_i^Q,KW_i^K,VW_i^V)  
  \end{split}
\end{equation}where $Concat$ represents to concatenate the results of heads, $W^O$, $W_i^Q$, $W_i^K$, $W_i^V$ are parameter matrices.
The Feed-Forward network performs further nonlinear transformations on the output of the Multi-Head attention to enhance the representation of the model.
The decoder block is also a stack of multiple decoder layers, 
but the decoder layer adds a Masked Multi-Head attention to prevent access to future information.

\section{Task Inference in Meta-RL}

In this section, we introduce in detail the architectural framework of the TIMRL, 
which utilizes the task inference model with GMM and transformer to generate task embeddings that aid in the training of policy networks.
In order to adapt to non-stationary and multi-task environments, 
we use GMM to extend the task representation and design the transformer-based recognition network to classify tasks. 
TIMRL decouples task inference and global policy learning and reduces the complexity of training. 
In addition, the recognition is individually trained using supervised learning in the task inference model to improve the accuracy of task recognition.

\begin{figure}[!t]
  \centering
  \includegraphics[width=3in]{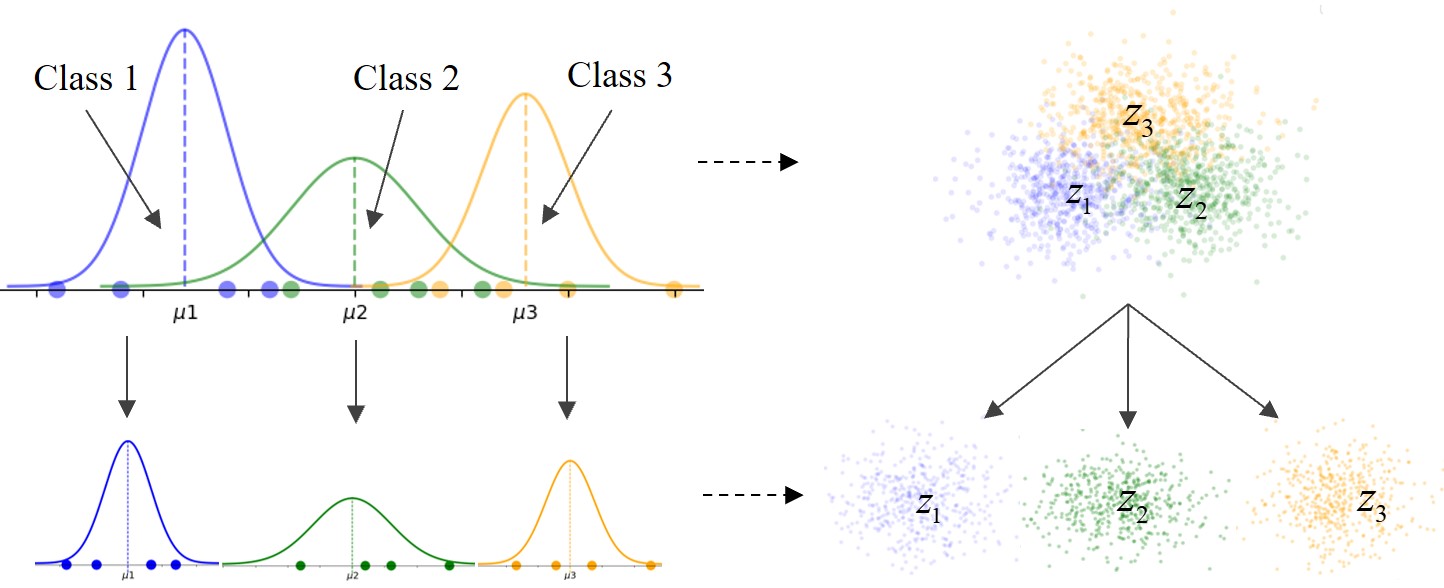}
  \caption{We use a GMM-based task inference model, 
  where the Gaussian component corresponding to the task is selected by the recognition network 
  and the task embedding $z$ is extracted from the Gaussian distribution.}
  \label{GMM}
\end{figure}

\subsection{Task Inference Model}

In traditional context-based meta-RL, the task representation capabilities of inferred networks based on a single Gaussian distribution are inadequate in some scenarios.
In contrast, our approach achieves the task inference model leveraging the framework of GMM, meaning the model needs to use multiple Gaussian distributions, 
as shown in Fig.~\ref{GMM}.
The latent variable $z$ sampled by the corresponding Gaussian component, 
which is selected by the transformer-based recognition network.

We construct recognition network based on transformer, which is aimed at learning how to accurately recognize tasks in the environment.
First, the pre-processing program is designed for pre-processing the input data $c$ of the recognition network,
which is the context during meta-training. 
The data cleansing is utilized to repair errors and dimensional inconsistencies in the data.
In addition, data pre-processing improves performance and reduces overfitting of the recognition network.

The transformer network completes task recognition based on the features of the input data and obtain recognition result $k_i$ for task $\mathcal{T}$:
\begin{equation}
  \label{deqn_ex8a}
  k_i=Trans(c_i^\mathcal{T}),c_i^\mathcal{T}\sim\mathcal{T}
\end{equation}where $Trans$ represents the recognition network, which contains the pre-processing program and the transformer network.

For the consideration of the efficiency of data utilization and the accuracy of the model, we utilize the supervised learning method to train the recognition network.
During the collecting experiences, the classes of context in each episode are labeled,
and placed into the replay buffer as $(s_i,a_i,r_i,s'_i)\rightarrow\hat{k_i}$.
Then the transformer network will recognize the class $k_i$ of context in meta-training.
We define the loss function:
\begin{equation}
  \label{deqn_ex9a}
  \begin{split}
  \mathcal{L}_{recognize}=MSE(\hat{k_i},k_i)
  =\frac{1}{n}\sum_{i=1}^n(\hat{k_i}-k_i)^2
  \end{split}
\end{equation}where $MSE$ represents the mean-square error function, and $n$ is the number of transition.

For different classifications of tasks, different Gaussian components are selected for task inference through the recognition network:
\begin{equation}
  \label{deqn_ex6a}
  \begin{aligned}  
  p_{k_i}(z|x_i)&=\left\{
  \begin{aligned}
  p_{1}(z|x_i),k_i=1 \\
  \dots,\dots\\
  p_{K}(z|x_i),k_i=K
  \end{aligned}
  \right.,\\
  z_i&\sim{p_{k_i}}(z|x_i)
  \end{aligned}
\end{equation}where $p_{k_i}(z|x_i)$ represents the $k_i$-th Gaussian distribution, 
$x_i$ is the input data, and GMM has $K$ components.

\subsection{Training Model}
In the task inference model, VAE achieves generative modeling by learning the distribution of latent space. 
Based on traditional VAE, there are two objectives to be minimized in the loss function, the reconstruction term and the regularization term. Considering the framework of GMM, we denote the loss function as:
\begin{equation}
  \label{deqn_ex6a}
  \begin{split}
  \mathcal{L}_{vae}=-\frac1K\sum_{k_i=1}^K\left(\mathbb{E}_{z\sim q_{\phi_{k_i}}(z|x_i)}[\log p_\theta(x_i|z)]\right.\\
  \left.-D_{KL}[q_{\phi_{k_i}}(z|x_i)||p_{k_i}(z)]\right)
  \end{split}
\end{equation}where $q_{\phi_{k_i}}(z|x_i)$, $ p_\theta(x_i|z)$ represent the encoder and decoder,
and $p_{k_i}(z)$ represents the prior of $z$.
To ensure proper propagation of the gradient during the sampling process, the reparameterization method is employed, $z=g(x,\epsilon),\epsilon\sim\mathcal{N}(0,I)$. 

We expect the output of decoder to attain the utmost possible value, thereby enhancing the likelihood of generating $x_i$.
This means that reconstruction term is employed to ensure that the output generated by the decoder closely resembles the input data as much as feasible.
The reconstruction loss is expressed using the mean-square error:
\begin{equation}
  \label{deqn_ex6a}
  \mathcal{L}_{recons}=\frac1m\sum_{i=1}^m\lVert x_i-{\hat{x_i}}\rVert^2
\end{equation}where $\hat{x_i}$ is the output decoder data, and $m$ is the number of data.

Unlike the reconstruction term, the regularization term avoids the disappearance of randomness and improves the diversity of outputs.
the regularization term is expressed using the KL divergence with the expectation that the distribution of latent space approximates the priori Gaussian distribution.
For the inference of various tasks in meta-RL, it is necessary to further refine the task inference model to clarify the boundaries between tasks.
We define different tasks corresponding to different prior distributions.
The regularization loss is expressed by the KL divergence between the Gaussian mixture and the priori distribution.
The $k_i$-th component corresponds to the parameters of the Gaussian distribution are $\mu_{k_i}$ and $\sigma_{k_i}^2$,
meanwhile, the parameters of the priori Gaussian distribution are $\hat{\mu_{k_i}}$ and $\hat{\sigma_{k_i}^2}$.
The regularization loss is defined as:
\begin{equation}
  \label{deqn_ex14a}
  \begin{split}
    \begin{aligned}
      \mathcal{L}_{regula}&{\operatorname*{=}}\sum_{k_i=1}^{K}{D}_{\mathrm{KL}}\left[\mathcal{N}(\mu_{k_i},\sigma_{k_i}^2)\parallel\mathcal{N}(\hat{\mu_{k_i}},\hat{\sigma_{k_i}^2})\right] \\
      &=-\frac12\sum_{k_i=1}^K\left(1-2\log\frac{\hat{\sigma_{k_i}^2}}{\sigma_{k_i}^2}-\frac{\sigma_{k_i}^2+(\mu_{k_i}-\hat{\mu_{k_i}})^2}{\hat{\sigma_{k_i}^2}}\right)
      \end{aligned}
  \end{split}
\end{equation}

\begin{figure*}[!t]
  \centering
  \includegraphics[width=5.5in]{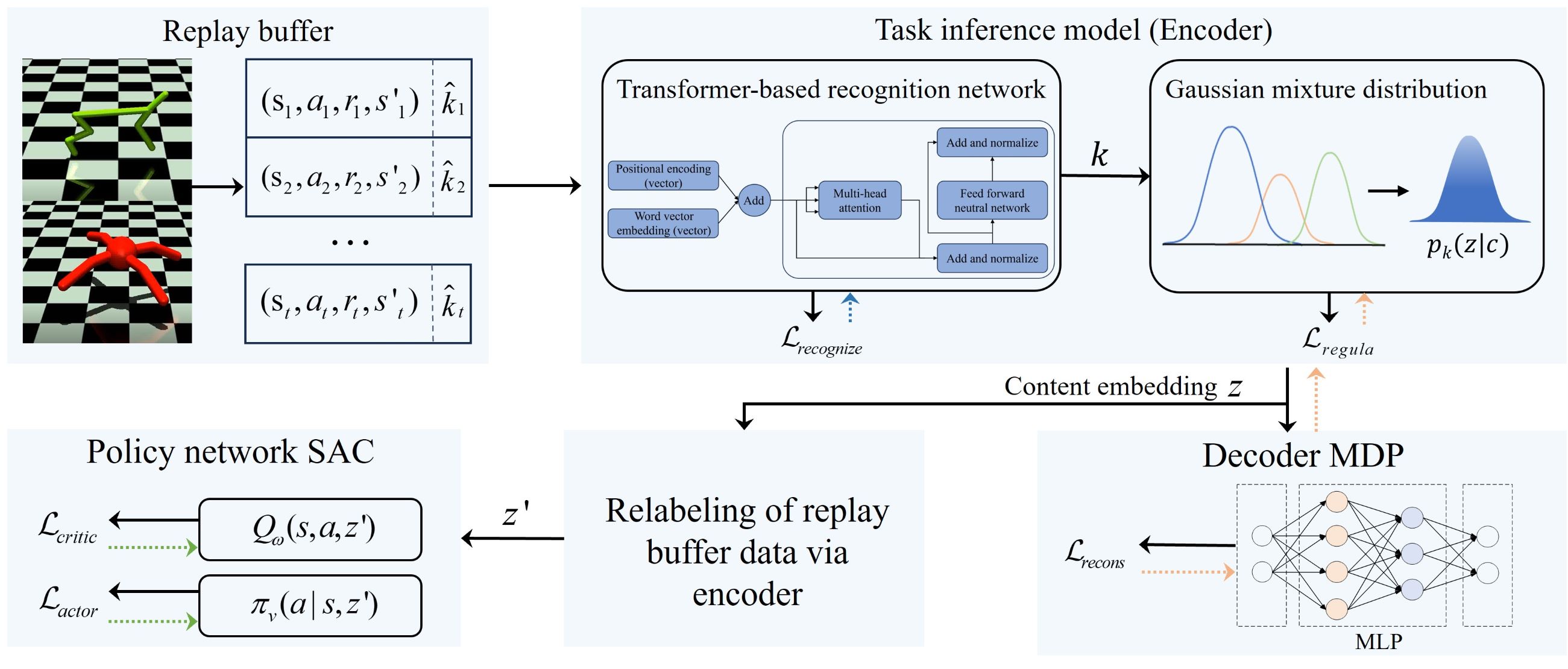}
  \caption{
    TIMRL primarily consists of a task inference model (encoder), a decoder, and a policy network (SAC). 
    We use GMM to construct task inference model and design a recognition network to classify tasks. 
    Based on the classification result, the corresponding Gaussian component is selected to generate the task embedding $z$. 
    The policy network conditioned on the task embedding is trained. The dashed arrows indicate the gradient backpropagation process.}
  \label{framework}
\end{figure*}

\subsection{TIMRL}

The framework of TIMRL is shown in Fig.~\ref{framework}, and the training process is summarised as pseudo-code in Algorithm~\ref{alg1}.
In meta-RL, when training tasks are given, we aim to leveraga the task inference model for accurately inferring the features of the tasks 
and training the policy to quickly solve new tasks. 
We focus on the task inference process and propose the task inference model based on the framework of GMM.
Utilizing the recognition network to classify various tasks and assigning each classification to a distinct Gaussian component enables
agent to learn optimal policy for each task more efficiently.

In TIMRL, we first design the MLP network for pre-processing context.
In the task inference phase, the inference model is utilized to obtain latent variable $z$ to guide the training of policy.
We can describe the probability of obtaining $z$ as the posterior $p(z|c)$, which is approximated by the encoder $q_\phi(z|c)$.
The encoder is constructed using the GMM architecture with $K$ components.
We utilize different components to fit the posterior corresponding to each classification of tasks.
The recognition network, as represented by $(\ref{deqn_ex8a})$, is responsible for recognizing the classification of each task and
updating it based on the current context and corresponding labels. 
The contribution of the encoder to the objective function of VAE corresponds to the regularization term.
Inspired by \cite{ref7}, we use the method of reconstructing MDP for decoding the latent variable. 
Specifically, the decoder with two specialized networks for reconstructing the next state and the reward. 
Input the tuple $(s,a,z)$ to the decoder network and the encoder network estimates the next state and reward.
This process can be described by the transition function $p(s_{t+1}|s_t,a_t,z_t)$ and $p(r_t|s_t,a_t,z_t)$.
So, the reconstruction term in the loss function of VAE can be rewritten as:
\begin{equation}
  \begin{split}
  \label{deqn_ex15a}
  \mathcal{L}_{recons}&=\mathcal{L}_{state}+\mathcal{L}_{reward} \\
  &=\|s_{t+1}-\hat{s}_{t+1}\|^2_2 +||r_{t}-\hat{r}_{t}||^{2}
  \end{split}
\end{equation}

\begin{algorithm}
  \caption{Meta-Training Procedure of TIMRL}
  \begin{algorithmic}[1]
  \REQUIRE $\text{Train task sets } \mathcal{T} = \{\mathcal{T}_i\}_{i=1}^T \sim p(\mathcal{T}) $
  \STATE $\text{Initialize encoder $\{q_{\phi_k}(z|x)\}_{k=1}^K$, decoder $p_\theta(\hat{x}|z)$ }$, actor $\pi_v(a|s,z)$, critic $Q_w(s,a,z)$ in SAC, recognition network $Trans$, and replay buffer $\mathcal{D}$
  \WHILE{not done}
  \FOR{train tasks ${\mathcal{T}_i} \sim \mathcal{T}$}
  \STATE Initialize context $c^{\mathcal{T}} = \{\}$
  \FOR{episode length}
  \STATE $k \sim Trans(c^{\mathcal{T}})$ and sample $z \sim q_{\phi_k}(z|c^{\mathcal{T}})$ 
  \STATE Gather $\hat{k}$ from ${\mathcal{T}_i}$, collect transition and add to $\mathcal{D}$
  \ENDFOR
  \ENDFOR
  \FOR{step in task inference training steps}
  \STATE Sample $c^{\mathcal{T}_t} \sim \mathcal{D}$
  \FOR{$k=1,...,K$}
  \STATE $z_t^{(k)} \sim q_{\phi_k}(z|c^{\mathcal{T}_t})$
  \STATE Calculate $\mathcal{L}_{regula}^{(k)}$ and $\mathcal{L}_{recons}^{(k)}$ with $(\ref{deqn_ex14a})$ and $(\ref{deqn_ex15a})$
  \ENDFOR
  \STATE Calculate $\mathcal{L}=\sum_{k=1}^K\left[\mathcal{L}_{recons}^{(k)}+\alpha\mathcal{L}_{regula}^{(k)}\right]$
  \STATE Update $(\{\phi_k\}_{k=1}^K, \theta)$ with gradients of $\mathcal{L}$
  \ENDFOR
  \FOR{step in recognition network training steps}
  \STATE Sample $c^{\mathcal{T}} \sim \mathcal{D}$ and $k \sim Trans(c^{\mathcal{T}})$
  \STATE Calculate $\mathcal{L}_{recognize}$ with $(\ref{deqn_ex9a})$
  \STATE Update $Trans$ with gradients of $\mathcal{L}_{recognize}$
  \ENDFOR
  \STATE Update $(v, w)$ with SAC algorithm
  \ENDWHILE

  \end{algorithmic}
  \label{alg1}
\end{algorithm}

According to $(\ref{deqn_ex14a})$ and $(\ref{deqn_ex15a})$, we propose to regularize the meta-training policy with the task distributions, 
leading to the following meta-training objective:
\begin{equation}
  \label{deqn_ex6a}
  \mathcal{L}=\mathcal{L}_{recons}+\alpha\mathcal{L}_{regula}
\end{equation}where $\alpha$ is the hyperparameter.

To accurately recognize the classification of the task, we decouple the GMM from the recognition network during training.
In the practical implementation, we train the recognition network using $(\ref{deqn_ex9a})$ after updating VAE in each epoch.
The input of the recognition network is sampled from the replay buffer that is shared concurrently with both the VAE and the policy learning.
This method not only improves efficiency but reduces complexity when training task inference model.

We use SAC for policy learning to realize off-policy training and the latent variable $z$ will be used as the auxiliary input to the SAC.
The objective function of SAC is:
\begin{equation}
  \begin{split}
    \mathcal{L}_{SAC}=\mathbb{E}_{\mathbf{s}_t\sim\mathcal{D}}\left[\mathbb{E}_{\mathbf{a}_t\sim\pi_\phi}
    \left[\alpha\log\left(\pi_v(\mathbf{a}_t|\mathbf{s}_t)\right)
    -Q_w(\mathbf{s}_t,\mathbf{a}_t)\right]\right]
  \end{split}
\end{equation}

\section{Experiments}
In this section, we introduce the non-stationary and multi-task environment
and evaluate the performance of TIMRL compared to state-of-the-art meta-RL algorithms.
In order to verify that our proposed recognition network and task inference model are effective, 
we performed ablation experiments on the relevant modules of our algorithm.

\begin{figure}[!t]
  \centering
  \includegraphics[width=3.5in]{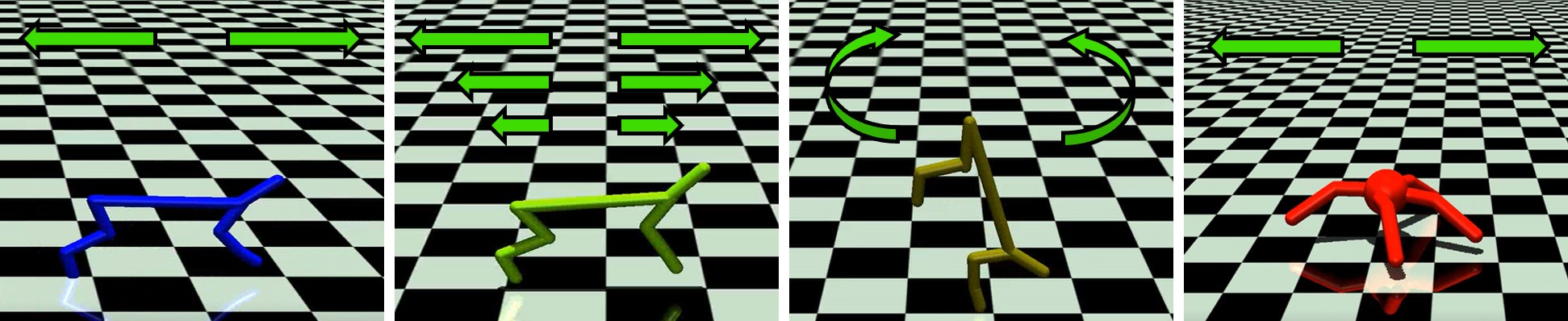}
  \vspace{-12pt}
  \caption{
    Four set of environments for algorithm evaluation. 
    From left to right the environments are Cheetah-Nonstat-Dir, Cheetah-Nonstat-Vel, Cheetah-Nonstat-Flipping, and Ant-Nonstat-Dir.}
    \label{3}
\end{figure}

\begin{figure*}[!t]
  \centering
  \captionsetup[subfigure]{labelformat=empty}
  \subfloat[]{\includegraphics[width=1.7in]{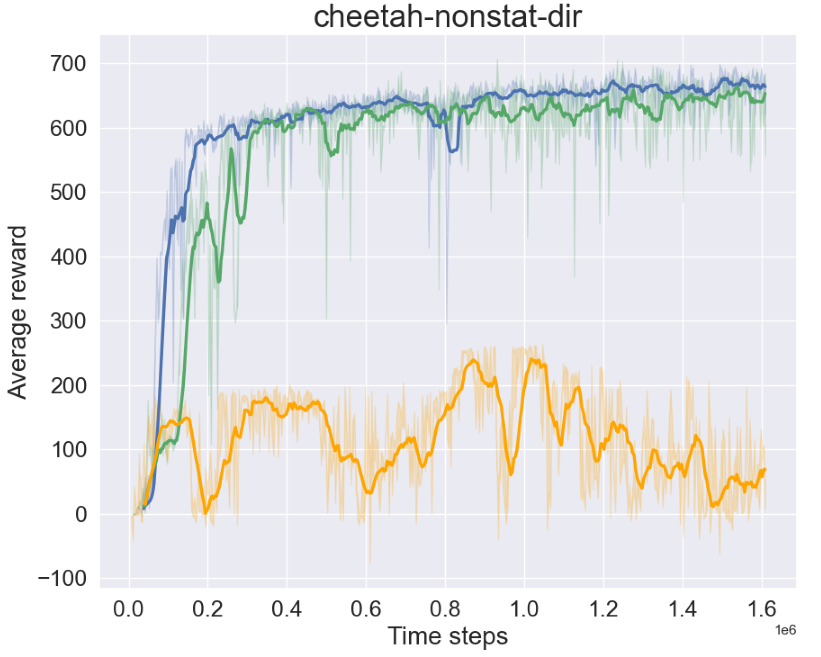}}
  \hfil
  \subfloat[]{\includegraphics[width=1.7in]{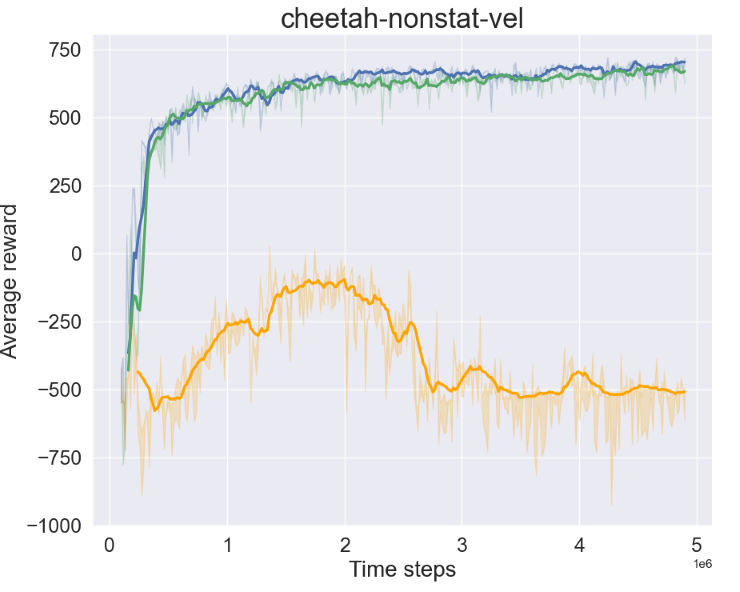}}
  \hfil
  \subfloat[]{\includegraphics[width=1.7in]{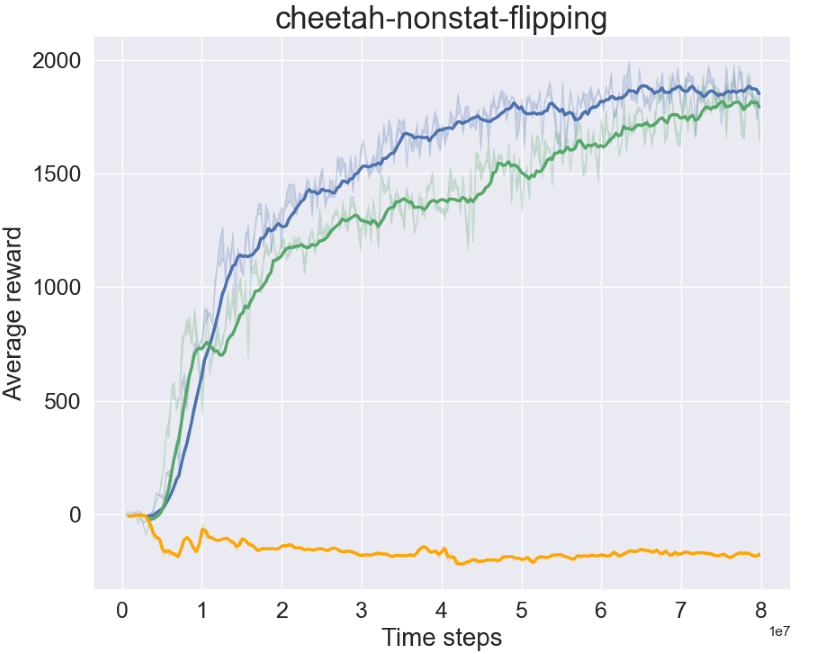}}
  \hfil
  \subfloat[]{\includegraphics[width=1.7in]{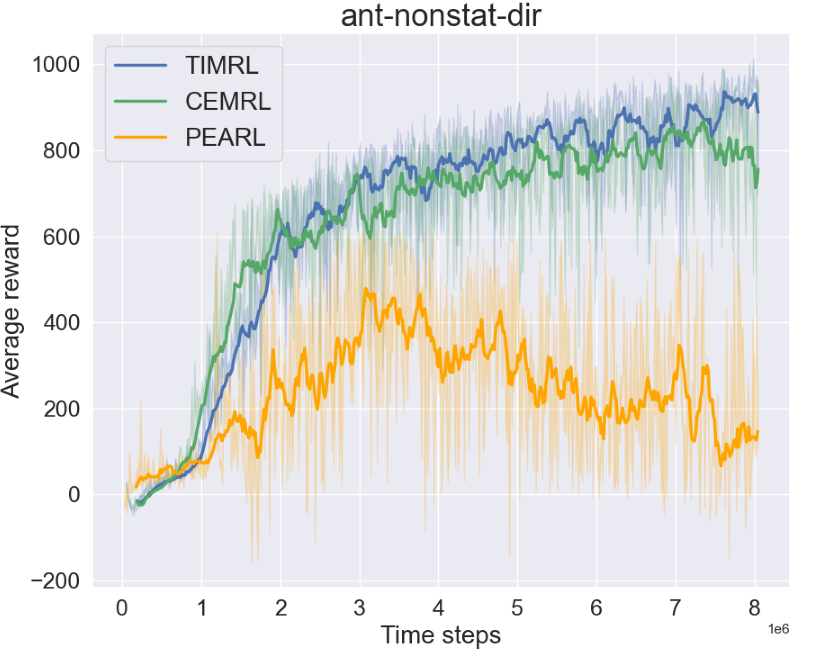}}
  \vspace{-15pt}
  \caption{Reward curves for the meta-testing task during training with the non-stationary environments.  
  These curves represent the performance of our algorithms in terms of asymptotic performance and sample efficiency in the above benchmark tasks. 
  The solid lines represent the asymptotic performance of each algorithm. 
  Non-stationary environments: Cheetah-Nonstat-Dir, Cheetah-Nonstat-Vel, Cheetah-Nonstat-Flipping, Ant-Nonstat-Dir.}
  \label{fig.4}
\end{figure*}

\begin{figure*}[!t]
  \centering
  \captionsetup[subfigure]{labelformat=empty}
  \subfloat[]{\includegraphics[width=1.6in]{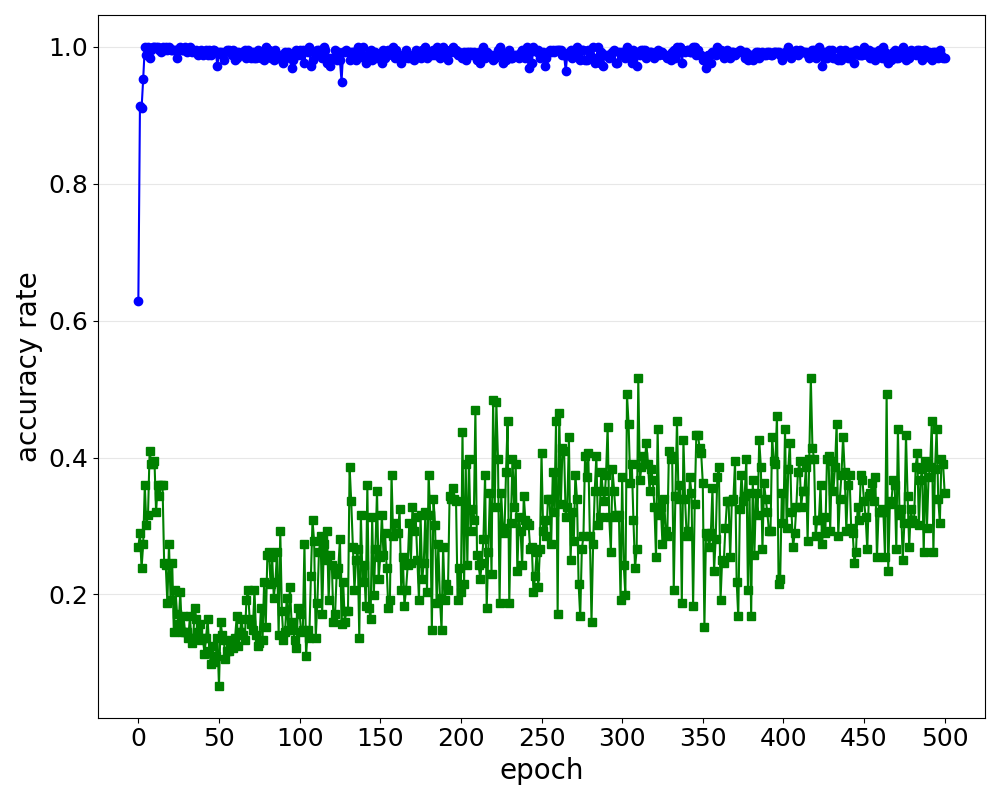}}
  \hfil
  \subfloat[]{\includegraphics[width=1.6in]{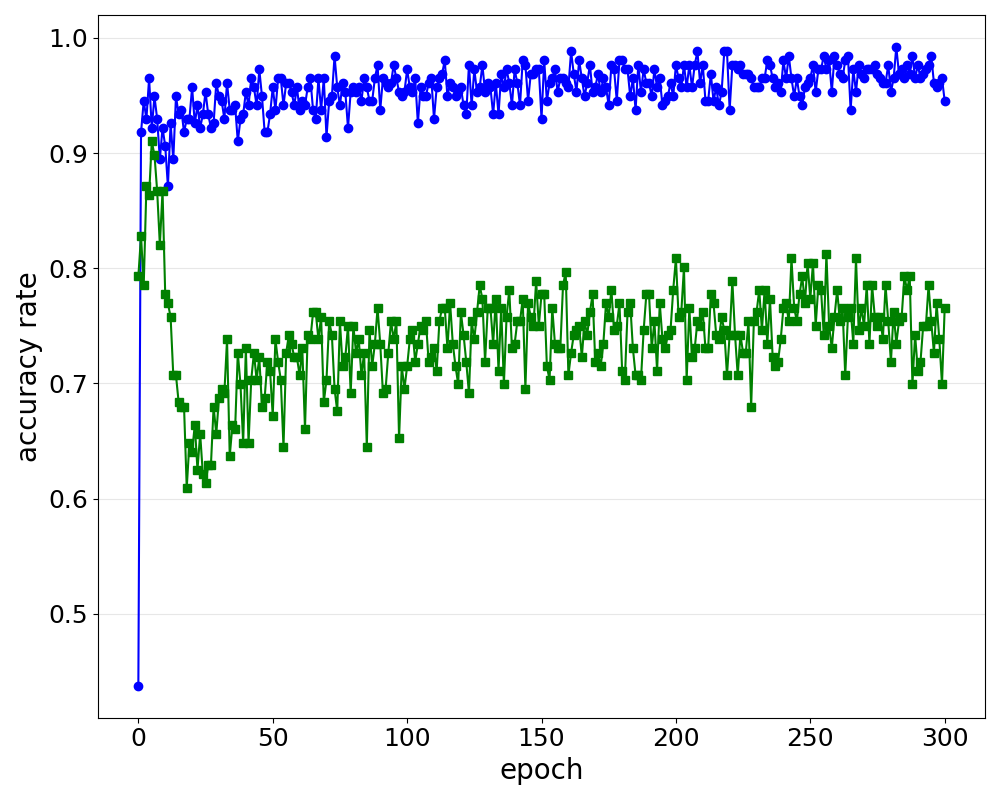}}
  \hfil
  \subfloat[]{\includegraphics[width=1.6in]{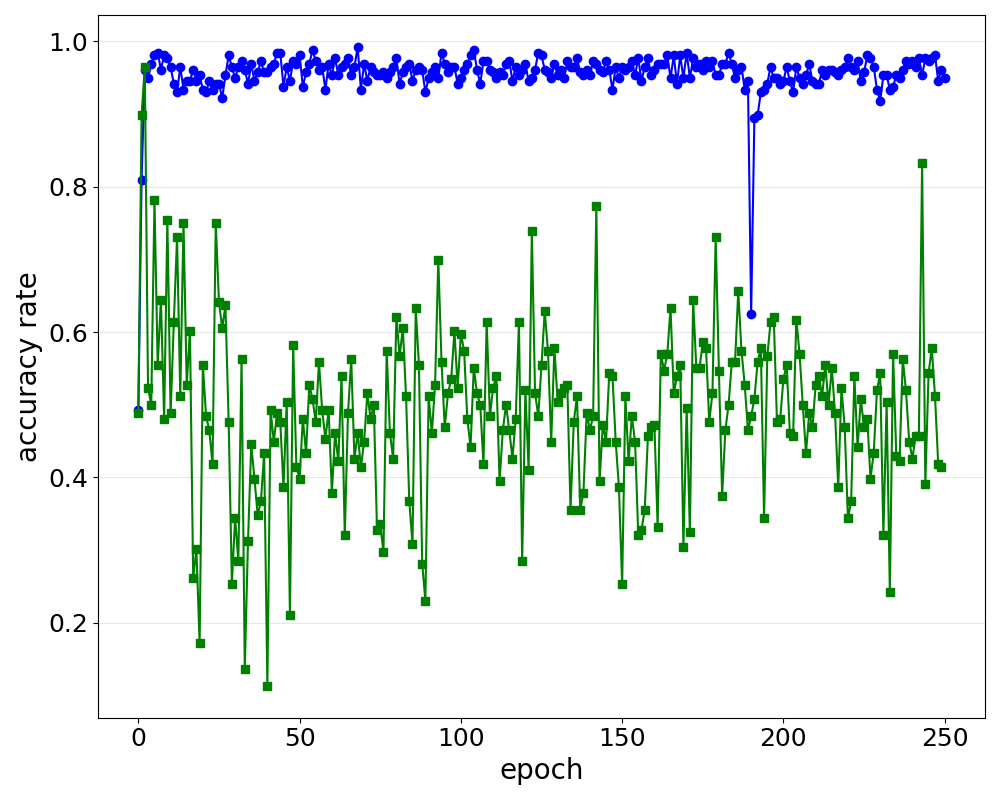}}
  \hfil
  \subfloat[]{\includegraphics[width=1.6in]{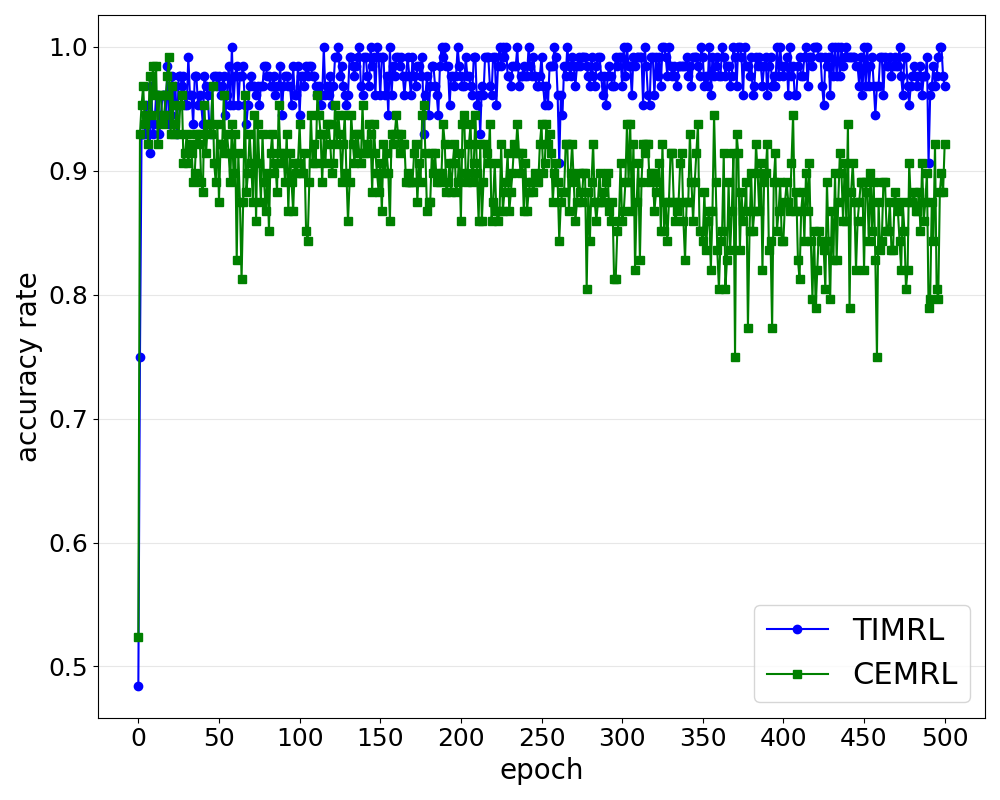}}
  \vspace{-15pt}
  \caption{Task recognition accuracy per epoch of the recognition network when trained in non-stationary environments corresponding to Fig.~\ref{fig.4}.}
  \label{fig.5}
\end{figure*}

\subsection{Experiments Setup}
We evaluate TIMRL by setting up non-stationary and multi-task environments with the MuJoCo benchmark.
Specifically, we extend the traditional stationary setup to build non-stationary environments, where the goal may change at any timestep within an episode,
that is to say, the MDP will change at any time.
Based on the above setup, we design the multi-task environment, i.e., completing multiple tasks in a single training process, 
to evaluate the adaptability of the proposed algorithm in the multi-task environment.
We provide a brief description of the setup of these environments as follows.

Non-stationary environment:

\begin{itemize}[leftmargin=2.5em]
  \item[1)] 
  Cheetah-Nonstat-Dir: Cheetah is required to randomly change direction of movement, forward or backward, in the episode, namely direction-following.
  \item[2)] 
  Cheetah-Nonstat-Vel: The speed of movement that Cheetah is required to achieve will change at any timestep in the episode, namely velocity-following.
  \item[3)]  
  Cheetah-Nonstat-Flipping: Cheetah is required to randomly change direction of flip, flipping forward or backward, in the episode, namely flip-following.
  \item[4)]
  Ant-Nonstat-Dir: Ant is required to randomly change direction of movement, forward or backward, in the episode.
  \end{itemize}

The four environments built are shown in Fig.~\ref{3}.

Multi-task environment:
\begin{itemize}[leftmargin=2.5em]
  \item[1)] 
  Cheetah-Dir/Goal: Cheetah is required to complete two classifications of tasks in the episode,
  including direction-following and goal-following. 
  The goal-following means the Cheetah is required to randomly change its target location in the episode.
  \item[2)]  
  Cheetah-Flipping/Jumping: Cheetah is required to complete two classifications of tasks in the episode, 
  including flipping-following task and jumping-following.
  The jumping-following means the Cheetah is required to jump randomly in the episode.
  \item[3)]  
  Cheetah-Vel/Goal/Dir: Cheetah is required to complete three classifications of tasks in the episode, 
  including velocity-following, direction-following and goal-following.
\end{itemize}

We set up a different number of training and testing tasks for each environment, and randomly select from these tasks when training the algorithm.
During the training process, we divide the samples in such a way that the percentage of training samples is 80\% and that of validation samples is 20\%.

\subsection{Performance}
We compare TIMRL with the classical context-based meta-RL algorithms PEARL\cite{ref5} and CEMRL\cite{ref7}, 
and complete training based on the setup of their proposed algorithms to obtain experimental results. 

$1)$ Non-stationary environment

We analyze the ability of algorithm to learn the behaviour of agent in new tasks when using TIMRL for training in non-stationary environment.
Since the tasks change randomly during an episode and rewards vary from one task to another, 
it is necessary to accurately identify the task classification in order to complete the tasks.
Fig.~\ref{fig.4} shows the rewards obtained by the learned policies of algorithms in the test tasks.

The result on the four task environments we set up shows that TIMRL is able to quickly adapt and learn in non-stationary environment, 
and its performance is superior to that of CEMRL in some of environments.
Reward curves indicate that our proposed task inference model can be adapted to the meta-RL learning process.
In contrast, PEARL does not perform well and is unable to learn the policy to complete the task in non-stationary environment,
as described above, we speculate that it is because the task inference model used by PEARL
is based on the Gaussian distribution and cannot well represent complex non-stationary tasks.

TIMRL exhibits good adaptability in non-stationary environments. 
In order to more accurately evaluate the performance of the recognition network in the task inference model, 
we further demonstrate the effectiveness of the recognition network. 
We propose an evaluation metric for recognition networks: the accuracy rate, i.e., the proportion of successfully recognizing the classification of tasks during training.
As show in Fig.~\ref{fig.5}, we demonstrates the results of task recognition during training.
In the first epoch of training, the task recognition results are poor due to the insufficient training of the recognition network, 
and the average accuracy rate of recognition reaches only about 60\%. 
As the training progresses, the task recognition accuracy rate converges steadily to 95\% after the 5 epoch. 
In contrast, CEMRL has a lower accuracy rate and is unable to accurately identify the classification of tasks.
The single Gaussian distribution is adopted by PEARL, which does not require classification, 
so its classification results are not presented in Fig.~\ref{fig.5}.
It is worth mentioning that although the accuracy rate did not reach 100\% in same environments, 
the classifications recognized by transformer belong to the tasks in each episode, 
and the impact of small errors on the results of subsequent task inferences could be negligible.

$2)$ Multi-task environment

\begin{figure*}[!t]
  \centering
  \captionsetup[subfigure]{labelformat=empty}
  \subfloat[]{\includegraphics[width=2.3in]{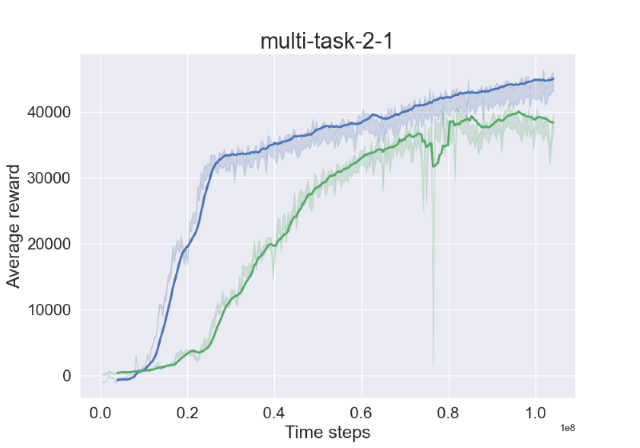}}
  \hfil
  \subfloat[]{\includegraphics[width=2.3in]{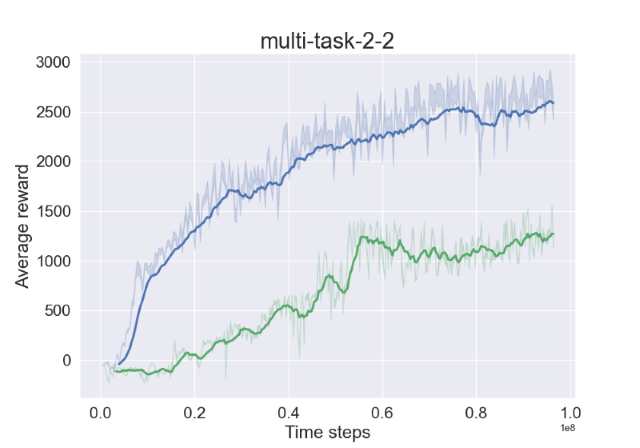}}
  \hfil
  \subfloat[]{\includegraphics[width=2.3in]{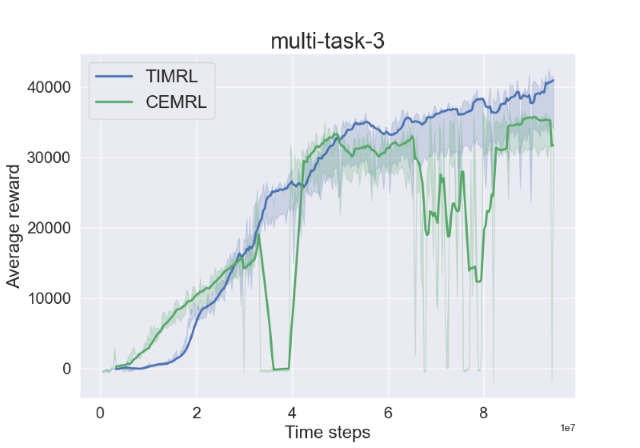}}
  \vspace{-15pt}
  \caption{Reward curves for the meta-testing task during training with the multi-task environments. 
  Multi-task environments: Cheetah-Dir/Goal, Cheetah-Flipping/Jumping and Cheetah-Vel/Goal/Dir.}
  \label{fig.6}
\end{figure*}

\begin{figure*}[!t]
  \centering
  \captionsetup[subfigure]{labelformat=empty}
  \hspace{-4mm}
  \subfloat[]{\includegraphics[width=2in]{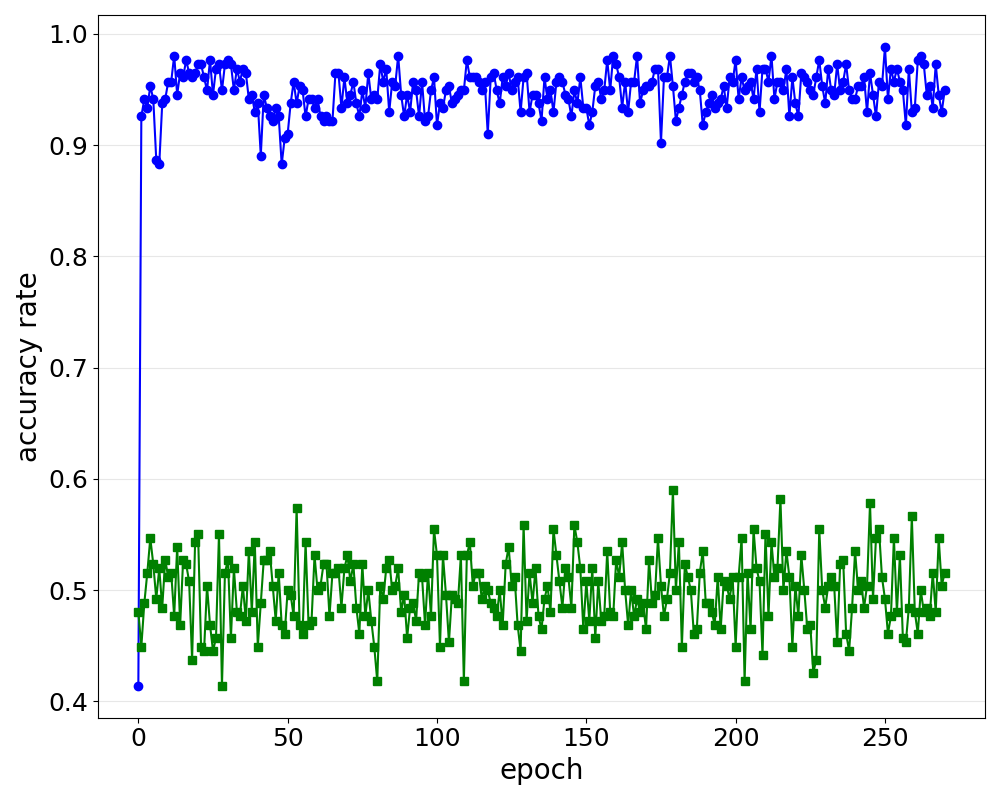}}
  \hfil
  \hspace{2mm}
  \subfloat[]{\includegraphics[width=2in]{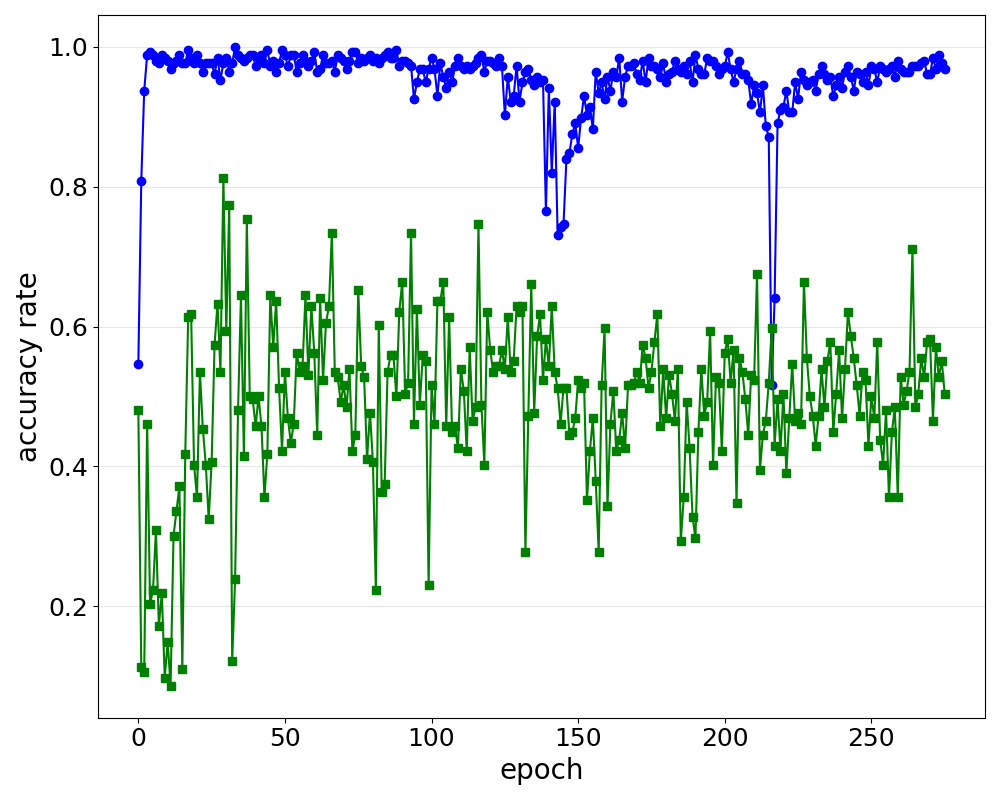}}
  \hfil
  \hspace{2mm}
  \subfloat[]{\includegraphics[width=2in]{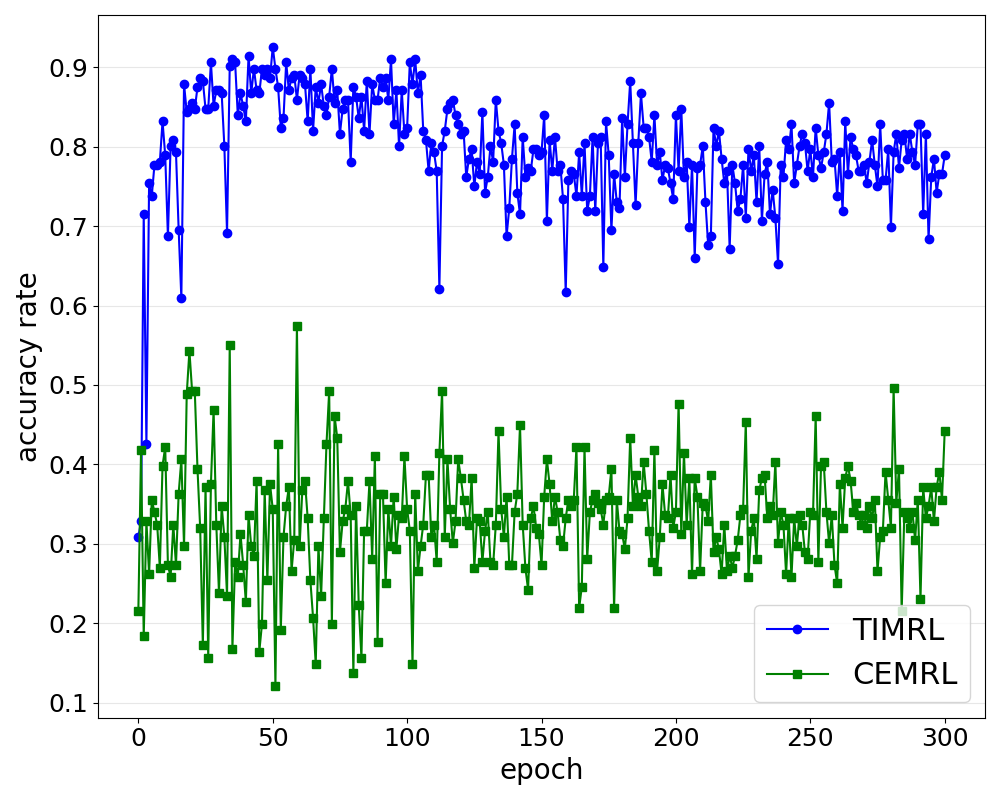}}
  \vspace{-15pt}
  \caption{Task recognition accuracy per epoch of the recognition network when trained in multi-task environments corresponding to Fig.~\ref{fig.6}.}
  \label{fig.7}
\end{figure*}

As mentioned above, the multi-task environment is more complex. 
It needs to complete tasks that include the direction-following task and the goal-following task and even more subtasks within one episode.

PEARL fails to adapt this environment and cannot perform policy learning, so we only compared the proposed algorithm with CEMRL.
Fig.~\ref{fig.6} shows the reward curves of the algorithms in the multi-task environments.
We can conclude that TIMRL can learn excellent policy more quickly in this challenging environment. 
This means that the task inference model we design is able to quickly recognize task classification and learn accurate task representations.

Similarly, we count the accuracy rate of task recognition in the multi-task environment as shown in Fig.~\ref{fig.7}.
At the end of epoch, the accuracy rate of task recognition are roughly converging to 95\% in Cheetah-Dir/Goal and Cheetah-Flipping/Jumping, 
and 80\% in Cheetah-Vel/Goal/Dir.
The result is better than that of CEMRL.
It can be concluded that the recognition network is able to accurately recognize the classification of tasks.
In Cheetah-Dir/Goal and Cheetah-Flipping/Jumping, tasks are divided into two classifications.
In Cheetah-Vel/Goal/Dir, task is divided into three classifications.
Unlike the results in the non-stationary environment, this classification is divided in terms of sub-tasks as a unit, 
such as the direction-following task and the goal-following task.

$3)$ Ablation Studies

\begin{figure}[!t]
  \centering
  \captionsetup[subfigure]{labelformat=empty}
  \subfloat[]{\includegraphics[width=1.8in]{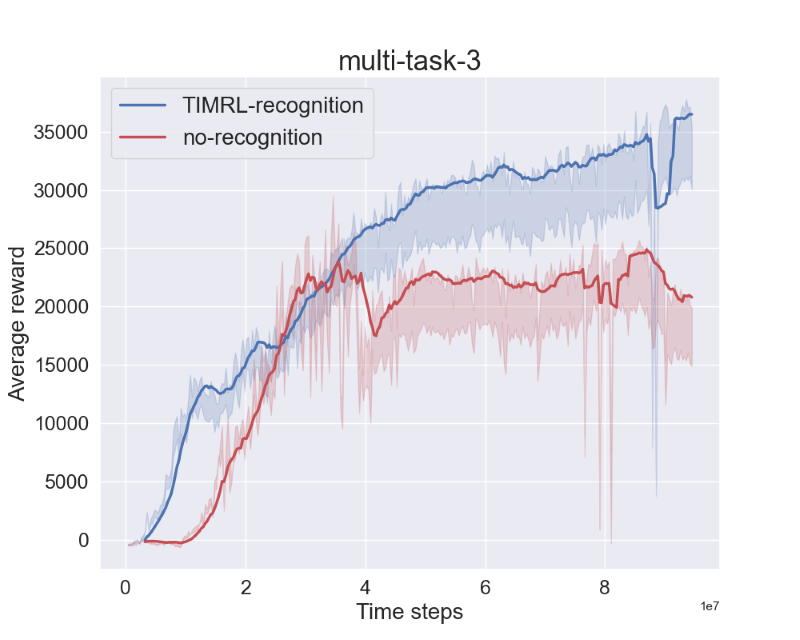}}
  \hfil
  \subfloat[]{\includegraphics[width=1.6in]{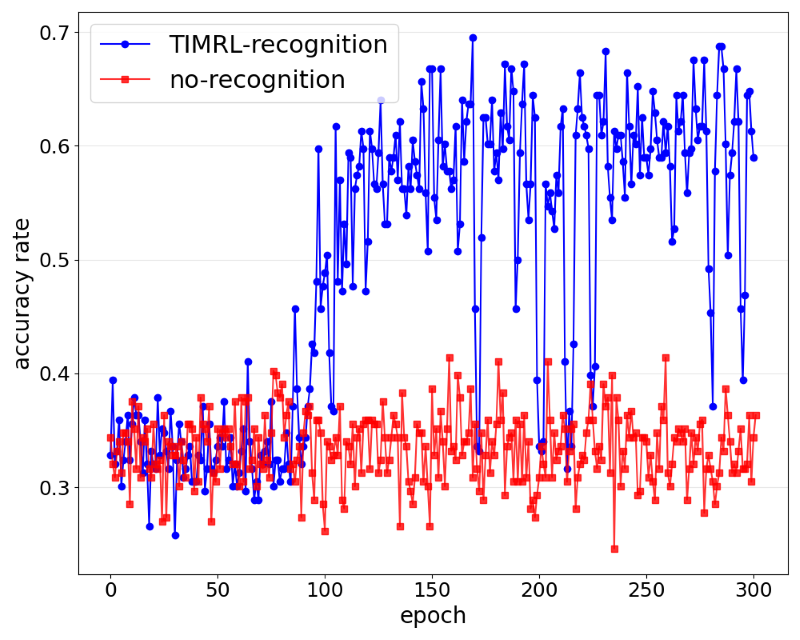}}
  \vspace{-15pt}
  \caption{Ablation experiment research. We present the reward curve for the ablation experiment (left).
  Task recognition accuracy of recognition networks during ablation experiments (right).}
  \label{fig.8}
\end{figure}

We explore the role of the proposed core task inference module, especially that of the task recognition network within it, through ablation studies.
In TIMRL, the recognition network helps the policy network to train faster and better by classifying and encoding the task.
In each epoch, the recognition network we have set up needs to be trained 10 times so that it can accurately recognize the tasks.
In order to verify the effectiveness of the recognition network, we make the results of the recognition network randomly selected. 
This will result in the algorithm cannot accurately determine the classification of task.
The results of the algorithm are then analysed to verify whether the accuracy of the task recognition network 
has a significant impact on the training of the algorithm, and to further illustrate the effectiveness of the task inference model.

Fig.~\ref{fig.8} shows the reward curve and the task recognition accuracy curve of the algorithm for ablation experiments,
from which it can be seen that inaccurate recognition results will lead to slow or even no convergence of the reward, 
as a proof of the effectiveness of our proposed task inference model.

\section{Conclusion}
In this paper, we proposed a novel meta-RL algorithm to adapt to non-stationary and multi-task environments.
A GMM-based task inference model was designed to encode tasks and assist in the training of the policy network. 
To improve the performance of the task inference model, 
a transformer-based task recognition network was designed to recognize tasks, 
and the recognition network was decoupled from the GMM to improve recognition accuracy and reduce training difficulty.
The TIMRL can effectively utilize contextual information to provide accurate task encoding for the training of policy networks 
and achieve SOTA performance.

\end{document}